\documentclass[runningheads]{llncs}
\usepackage{graphicx}
\usepackage{comment}
\usepackage{amsmath,amssymb} 
\usepackage{color}
\usepackage{url}
\usepackage{paralist}
\usepackage{subfig}
\usepackage{booktabs}
\usepackage{makecell}
\usepackage{gensymb}
\usepackage{pifont}
\usepackage{paralist}
\usepackage{balance}
\usepackage[pagebackref=true,breaklinks=true,letterpaper=true,colorlinks,bookmarks=false,citecolor=blue,linkcolor=blue]{hyperref}

\usepackage[width=122mm,left=12mm,paperwidth=146mm,height=193mm,top=12mm,paperheight=217mm]{geometry}

\newcommand{\cmark}{\ding{51}}%
\def\eg{e.g.,~}               
\def\ie{i.e.,~}               
\def\etc{etc}                 

\newlength\paramargin
\newlength\figmargin
\newlength\subfigmargin
\newlength\secmargin
\newlength\subsecmargin
\newlength\tabmargin
\newlength\eqmargin

\newcommand{\subsecref}[1]{Section~\ref{subsec:#1}}
\newcommand{\figref}[1]{Figure~\ref{figure:#1}} 
\newcommand{\tabref}[1]{Table~\ref{tab:#1}}
\newcommand{\eqnref}[1]{\eqref{eq:#1}}

\long\def\ignorethis#1{}

\newcommand{\tb}[1]{\textbf{#1}}

\begin{document}
\pagestyle{headings}
\mainmatter
\def\ECCVSubNumber{574}  

\title{Neural Design Network: Graphic Layout Generation with Constraints} 

\author{
Hsin-Ying Lee\thanks{Work done during their internship at Google Research.}$^2$,
Lu Jiang$^1$,
Irfan Essa$^{1,4}$,\\
Phuong B Le$^1$, 
Haifeng Gong$^1$,
Ming-Hsuan Yang$^{1,2,3}$,
Weilong Yang$^1$
}
\authorrunning{H.-Y. Lee et al.}
%
\institute{$^1$Google Research\hspace{14pt}$^2$University of California, Merced\hspace{14pt}\\$^3$Yonsei University\hspace{14pt}$^4$Georgia Institute of Technology}

\maketitle

\begin{figure}[th]
\centering
\includegraphics[width=\linewidth]{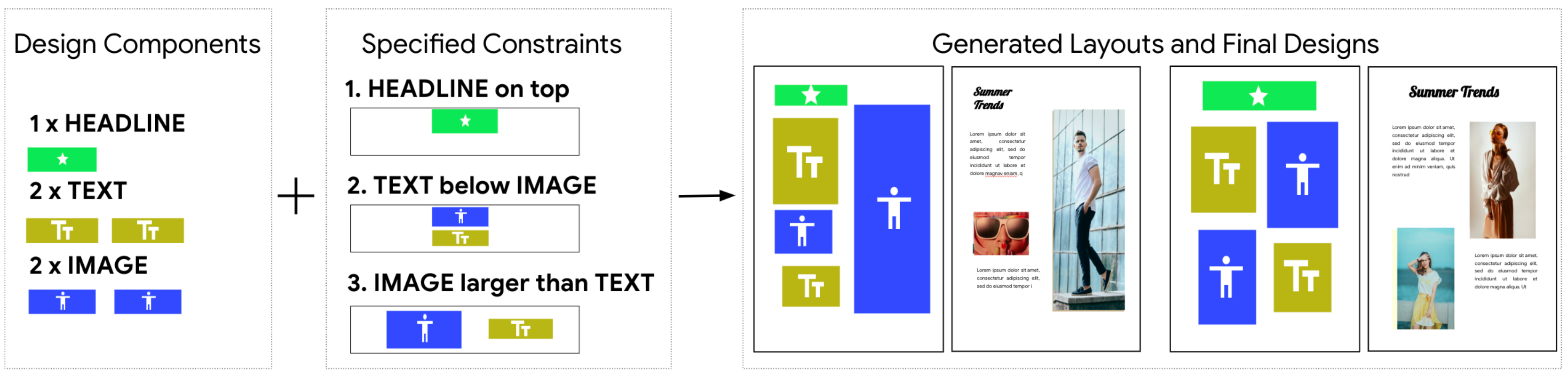}
    \captionof{figure}{
    \textbf{Graphic layout generation with user constraints.}
    We present realistic use cases of the proposed model.
    Given the desired components and partial user-specified constraints among them, our model can generate layouts following these constraints.
    We also present example designs constructed based on the generated layouts.
    }
    \label{figure:teaser}
    \vspace{-5mm}
\end{figure}

\begin{abstract}
Graphic design is essential for visual communication with layouts being fundamental to composing attractive designs. Layout generation differs from pixel-level image synthesis and is unique in terms of the requirement of mutual relations among the desired components. We propose a method for design layout generation that can satisfy user-specified constraints. The proposed neural design network (NDN) consists of three modules. The first module predicts a graph with complete relations from a graph with user-specified relations. The second module generates a layout from the predicted graph. Finally, the third module fine-tunes the predicted layout.
Quantitative and qualitative experiments demonstrate that the generated layouts are visually similar to real design layouts.  We also construct real designs based on predicted layouts for a better understanding of the visual quality. Finally, we demonstrate a practical application on layout recommendation.
\end{abstract}

\section{Introduction}
\vspace{\secmargin}
\label{sec:intro}
Graphic design surrounds us on a daily basis, from image advertisements, movie posters, and book covers to more functional presentation slides, websites, and mobile applications. Graphic design is a process of using text, images, and symbols to visually convey messages. Even for experienced graphic designers, the design process is iterative and time-consuming with many false starts and dead ends. This is further exacerbated by the proliferation of platforms and users with significantly different visual requirements and desires.

In graphic design, layout -- the placement and sizing of components (\eg title, image, logo, banner, \etc.) -- plays a significant role in dictating the flow of the viewer's attention and, therefore, the order by which the information is received. Creating an effective layout requires understanding and balancing the complex and interdependent relationships amongst all of the visible components. Variations in the layout change the hierarchy and narrative of the message.

In this work, we focus on the layout generation problem that places components based on the component attributes, relationships among components, and user-specified constraints. \figref{teaser} illustrates examples where users specify a collection of assets and constraints, then the model would generate a design layout that satisfies all input constraints, while remaining visually appealing.  
Generative models have seen a success in rendering realistic natural images~\cite{goodfellow2014generative,karras2019style,razavi2019generating}. However, learning-based graphic layout generation remains less explored. Existing studies tackle layout generation based on templates~\cite{damera2011probabilistic,hurst2009review} or heuristic rules~\cite{o2014learning}, and more recently using learning-based generation methods~\cite{jyothi2019layoutvae,li2019layoutgan,zheng2019contentaware}. However, these approaches are limited in handling relationships among components. High-level concepts such as mutual relationships of components in a layout are less likely to be captured well with conventional generative models in pixel space. Moreover, the use of generative models to account for user preferences and constraints is non-trivial. Therefore, effective feature representations and learning approaches for graphic layout generation remain challenging.

In this work, we introduce neural design network (NDN), a new approach of synthesizing a graphic design layout given a set of components with user-specified attributes and constraints. We employ directional graphs as our feature representation for components and constraints since the attributes of components (node) and relations among components (edge) can be naturally encoded in a graph. NDN takes as inputs a graph constructed by desired components as well as user-specified constraints, and then outputs a layout where bounding boxes of all components are predicted. 
NDN consists of three modules. First, the \textit{relation prediction} module takes as input a graph with partial edges, representing components and user-specified constraints, and infers a graph with complete relationships among components. Second, in the \textit{layout generation} module, the model predicts bounding boxes for components in the complete graph in an iterative manner. Finally, in the \textit{refinement} module, the model further fine-tunes the bounding boxes to improve the alignment and visual quality.

We evaluate the proposed method qualitatively and quantitatively on three datasets under various metrics to analyze the visual quality. The three experimental datasets are RICO~\cite{rico1,rico2}, Magazine~\cite{zheng2019contentaware}, and an image banner advertisement dataset collected in this work. These datasets reasonably cover several typical applications of layout design with common components such as images, texts, buttons, toolbars and relations such as above, larger, around, etc. We construct real designs based on the generated layouts to assess the quality. We also demonstrate the efficacy of the proposed model by introducing a practical layout recommendation application. 

To summarize, we make the following contributions in this work:
\begin{compactitem}
\itemsep 0em 
  \item  We propose a new approach that can generate high-quality design layouts for a set of desired components and user-specified constraints. 
  \item  We validate that our method performs favorably against existing models in terms of realism, alignment, and visual quality on three datasets.
  \item We demonstrate real use cases that construct designs from generated layouts and a layout recommendation application. 
  Furthermore, we collect a real-world advertisement layout dataset to broaden the variety of existing layout benchmarks.
\end{compactitem}
\section{Related Work}
\label{sec:related}
\vspace{\secmargin}
\textbf{Natural scene layout generation.}
Layout is often used as the intermediate representation in the image generation task conditioned on text~\cite{gupta2018imagine,hong2018inferring,tan2018text2scene} or scene graph~\cite{johnson2018image}.
Instead of directly learning the mapping from the source domain (\eg text and scene graph) to the image domain, these methods model the operation as a two-stage framework.
They first predict layouts conditioned on the input sources, and then generate images based on the predicted layouts.
Recently, Jyothi {\em et al.} propose the LayoutVAE~\cite{jyothi2019layoutvae}, which is a generative framework that can synthesize scene layout given a set of labels.
However, a graphic design layout has several fundamental differences to a natural scene layout.
The demands for relationship and alignment among components are strict in graphic design.
A few pixels offsets of components can either cause a difference in visual experience or even ruin the whole design.
The graphic design layout does not only need to look realistic but also needs to consider the aesthetic perspective.

\begin{figure*}[t]
    \centering
    \includegraphics[width=\linewidth]{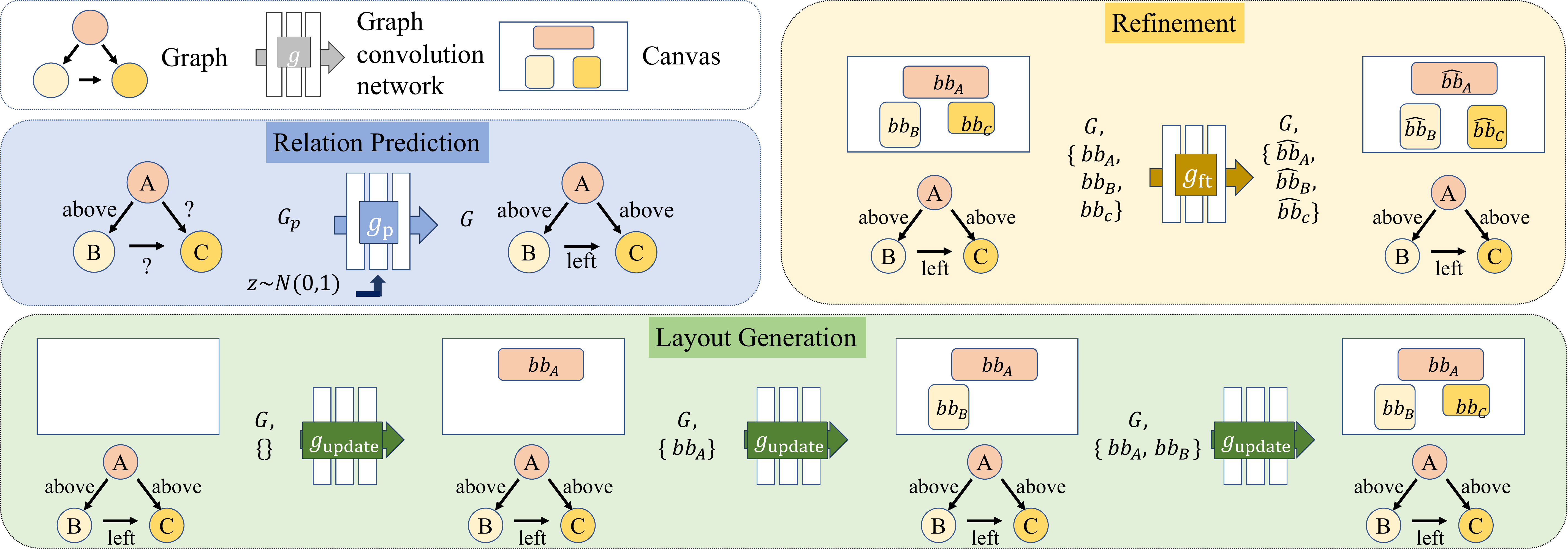}
    \caption{\textbf{Framework illustration.}
    Neural design network consists of three modules: relation prediction, bounding box prediction, and refinement.
    We illustrate the process with a three-component example.
    In \textbf{relation prediction} module, the model takes as inputs a graph with partial relations along with a latent vector (encoded from the graph with complete relations during training, sampled from prior during testing), and outputs a graph with complete relations.
    Only the graph with location relations is shown in the figure for brevity.
    In \textbf{layout generation} module, the model takes a graph with complete relations as inputs, and predicts the bounding boxes of components in an iterative manner.
    In \textbf{refinement} module, the model further fine-tune the layout.
    }
    \label{figure:framework}
    \vspace{\figmargin}
\end{figure*}

\vspace{\paramargin}
\noindent\textbf{Graphic design layout generation.}
Early work on design layout or document layout mostly relies on templates~\cite{damera2011probabilistic,hurst2009review}, exemplars~\cite{kumar2011bricolage}, or heuristic design rules~\cite{o2014learning,tabata2019automatic}.
These methods rely on predefined templates and heuristic rules, for which professional knowledge is required.
Therefore, they are limited in capturing complex design distributions.
Other work leverages saliency maps~\cite{bylinskii2017learning} and attention mechanisms~\cite{pang2016directing} to capture the visual importance of graphic designs and to trace the user's attention.
Recently, generative models are applied to graphic design layout generation~\cite{li2019layoutgan,zheng2019contentaware}.
The LayoutGAN model~\cite{li2019layoutgan} can generate layouts consisting of graphic elements like rectangles and triangles.
However, the LayoutGAN model generates layout from input noises and fails to handle layout generation given a set of components with specified attributes, which is the common setting in graphic design.
The Layout Generative Network~\cite{zheng2019contentaware} is a content-aware layout generation framework that can render layouts conditioned on attributes of components.
While the goals are similar, the conventional GAN-based framework cannot explicitly model relationships among components and user-specified constraints.

\vspace{\paramargin}
\noindent\textbf{Graph neural networks in vision.}
Graph Neural Networks (GNNs)~\cite{goller1996learning,gori2005new,scarselli2008graph} aim to model dependence among nodes in a graph via message passing.
GNNs are useful for data that can be formulated in a graph data structure.
Recently, GNNs and related models have been applied to classification~\cite{kipf2016semi}, scene graph~\cite{cheng2020rsegvae,johnson2018image,li2019controllable,tseng2020retrievegan,yang2018graph}, motion modeling~\cite{jain2016structural}, and molecular property prediction~\cite{duvenaud2015convolutional,jin2018learning}, to name a few.
In this work, we model a design layout as a graph and apply GNNs to capture the dependency among components.
\vspace{\secmargin}
\section{Graphic Layout Generation}
\label{sec:method}
\vspace{\secmargin}
Our goal is to generate design layouts given a set of design components with user-specified constraints.
For example, in image ads creation, the designers can input the constraints such as ``logo at bottom-middle of canvas'', ``call-to-action button of size (100px, 500px)'', ``call-to-action-button is below logo'', etc.
The goal is to synthesize a set of design layouts that satisfy both the user-specified constraints as well as common rules in image ads layouts.
Unlike layout templates, these layouts are dynamically created and can serve as inspirations for designers.

We introduce the neural design network using graph neural network and conditional variational auto-encoder (VAE)~\cite{kingma2013auto,rezende2014stochastic} with the goal of capturing better representations of design layouts. 
\figref{framework} illustrates the process of generating a three-component design with the proposed neural design network.
In the rest of this section, we first describe the problem overview in \subsecref{formulation}.
Then we detail three modules in NDN: the relation prediction (\subsecref{stage1}) and layout generation modules (\subsecref{stage2}), and  refinement module (\subsecref{stage3}).
%

\subsection{Problem Overview}
\label{subsec:formulation}
\vspace{\subsecmargin}
The inputs to our network are a set of design components and user-specified constraints.
We model the inputs as a graph, where each design component is a node and their relationships are edges. 
In this paper, we study two common relationships between design components: \emph{location} and \emph{size}.

Define $G=\{G_{\mathrm{loc}}, G_{\mathrm{size}}\} = (O, E_{\mathrm{loc}}, E_{\mathrm{size}})$, where $O=\{o_0, o_1,...,o_n\}$ is a set of $n$ components with each $o_i\in \mathcal{C}$ coming from a set of categories $\mathcal{C}$. 
We use $o_0$ to denote the \emph{canvas} that is fixed in both location and size, and $o_i$ to denote other design components that need to be placed on the \emph{canvas}, such as \emph{logo}, \emph{button}.
$E_{\mathrm{loc}}=\{l_1\,...,l_{m_l}\}$ and $E_{\mathrm{size}}=\{s_1\,...,s_{m_s}\}$ are sets of directed edges with $l_k=(o_i,r_l,o_j)$ and $s_k=(o_i,r_s,o_j)$, where $r_l\in \mathcal{R}_{\mathrm{loc}}$ and $r_s\in\mathcal{R}_{\mathrm{size}}$. 
Here, $\mathcal{R}_{\mathrm{size}}$ specifies the relative size of the component, such as \emph{smaller} or \emph{bigger}, and
$r_l$ can be \emph{left}, \emph{right}, \emph{above}, \emph{below}, \emph{upper-left}, \emph{lower-left}, etc. In addition, if anchoring on the \emph{canvas} $o_0$, we extend the $\mathcal{R}_{\mathrm{loc}}$ to capture the location that is relative to the \emph{canvas}, \emph{e.g.}, upper-left of the canvas.

Furthermore, in reality, designers often do not specify all the constraints. This results in an input graph with missing edges. \figref{framework} shows an example of a three-component design with only one specified constraint  ``($A$, above, $B$)'' and several unknown relations  ``$?$''.
To this end, we augment $\mathcal{R}_{\mathrm{loc}}$ and $\mathcal{R}_{\mathrm{size}}$ to include an additional \textit{unknown} category, and represent graphs that contain unknown size or location relations as $G^\mathrm{p}_{\mathrm{size}}$ and $G^\mathrm{p}_{\mathrm{loc}}$, respectively, to indicate they are the partial graphs. In \subsecref{stage1}, we describe how to predict the unknown relations in the partial graphs. 

Finally, we denote the output layout of the neural design network as a set of bounding boxes $\{bb_1, \, ..., bb_{|O|}\}$, where $bb_i=\{x_i, y_i, w_i, h_i\}$ represents the location and shape. 

In all modules, we apply the graph convolutional networks on graphs.
The graph convolutional networks take as the input the features of nodes and edges, and outputs updated features.
The input features can be one-hot vectors representing the categories or any embedded representations.

More implementation details can be found in the supplementary material. 
%

\subsection{Relation Prediction}
\label{subsec:stage1}
\vspace{\subsecmargin}
%
%
%
In this module, we aim to infer the unknown relations in the user-specified constraints. \figref{framework} shows an example where a three-component graph is given and we need to predict the missing relations between $A$, $B$, and $C$.
For brevity, we denote the graphs with complete relations as $G$, and the graphs with partial relations as $G^{\mathrm{p}}$, which can be either $G^{\mathrm{p}}_{\mathrm{size}}$ or $G^{\mathrm{p}}_{\mathrm{loc}}$.
Note that since the augmented relations include the \textit{unknown} category, both $G^{\mathrm{p}}$ and $G$ are complete graphs in practice.
We also use $e_i$ to refer to either $l_i$ or $s_i$ depending on the context.

We model the prediction process as a paired graph-to-graph translation task: from $G^{\mathrm{p}}$ to $G$.
Since the translation is multimodal, \ie a graph with partial relations can be translated to many possible graphs with complete relations.
Therefore, we adopt a similar framework to the multimodal image-to-image translation~\cite{zhu2017toward} and treat $G^{\mathrm{p}}$ as the source domain and $G$ as the target domain. 
Similar to~\cite{zhu2017toward}, the translation is a conditional generation process that maps the source graph, along with a latent code, to the target graph.
The latent code is encoded from the corresponding target graph $G$ to achieve reconstruction during training, and is sampled from a prior during testing.
The conditional translation encoding process is modeled as:
\begin{equation}
    \begin{aligned}
    z &= g_c(G) && z\in \mathcal{Z},\\
    \{h_i\} &= g_p(G^{\mathrm{p}}, z) &&i=1,...,|\tilde{E}|,\\
    \{\hat{e}_i\} &= h_{\mathrm{pred}}(\{h_i\}) &&  i=1,...,|E|,
    \end{aligned}
\end{equation}
where $g_{c}$ and $g_{p}$ are graph convolutional networks, and $h_{\mathrm{pred}}$ is a relation predictor. 
In addition, $\tilde{E}$ is the set of edges in the target graph. 
Note that $|\tilde{E}|=|E|$ since the graph is a complete graph.

The model is trained with the reconstruction loss $L_{\mathrm{cls}}=\mathrm{CE}(\{\hat{e}_i\}, \{e_i\})$ on the relation categories, where the $\mathrm{CE}$ indicates cross-entropy function, and a KL loss on the encoded latent vectors to facilitate sampling at inference time:  $L_{\mathrm{KL_1}}= \mathbb{E}[D_{\mathrm{KL}}((z)\| \mathcal{N}(0,1))]$, where $D_{\mathrm{KL}}(p\|q)=-\int{p(z)\log{\frac{p(z)}{q(z)}}\mathrm{d}z}$.
The objective of the relation prediction module is:
\begin{equation}
L_{\mathrm{rel}}=\lambda_{\mathrm{cls}}L_{\mathrm{cls}}+\lambda_{\mathrm{KL_1}}L_{\mathrm{KL_1}}.
\end{equation}
The reconstruction loss captures the knowledge that the predicted relations should agree with the existing relations in $G^{\mathrm{p}}$, and fill in any missing edge with the most likely relation discovered in the training data.

\subsection{Layout Generation}
\label{subsec:stage2}
\vspace{\subsecmargin}
Given a graph with complete relations, this module aims to generate the design layout by predicting the bounding boxes for all nodes in the graph.
%

Let $G$ be the graph with complete relations constructed from $G_{\mathrm{size}}$ and $G_{\mathrm{loc}}$, the output of the relation prediction module.
We model the generation process using a graph-based iterative conditional VAE model.
We first obtain the features of each component by
\begin{equation}
    \{f_i\}_{i=1\sim |O|} = g_\mathrm{enc}(G),
\end{equation}
where $g_\mathrm{enc}$ is a graph convolutional network.
These features capture the relative relations among all components.
We then predict bounding boxes in an iterative manner starting from an empty canvas (\ie all bounding boxes are unknown).
As shown in \figref{framework}, the prediction of each bounding box is conditioned on the initial features as well as the current canvas, \ie predicted bounding boxes from previous iterations.
At iteration $k$, the condition can be modeled as:
\begin{equation}
\begin{aligned}
    t_k &= ( \{f_i\}_{i=1\sim |O|}, \{bb_i\}_{i=1\sim k-1}), \\
    c_k &= g_{\mathrm{update}}(t_k),\\
     \end{aligned}
\end{equation}
where $g_{\mathrm{update}}$ is another graph convolutional network. $t_k$ is a tuple of features and current canvas at iteration $k$, and $c_k$ is a vector.
Then we apply conditional VAE on the current bounding box $bb_k$ conditioned on $c_k$.
\begin{equation}
    \label{eq:stage2}
    \begin{aligned}
         z &= h_{bb}^{\mathrm{enc}}(bb_k, c_k),\\
         \hat{bb}_k &= h_{bb}^{\mathrm{dec}}(z, c_k),
    \end{aligned}
\end{equation}
where $h_{bb}^{\mathrm{enc}}$ and $h_{bb}^{\mathrm{dec}}$ represent encoders and decoders consisting of fully connected layers.
%
%
We train the model with conventional VAE loss: a reconstruction loss $L_{\mathrm{recon}}= \displaystyle\sum_{i=1}^{|O|}\lVert \hat{bb}_i - bb_i  \lVert_{1}$ and a KL loss $L_{\mathrm{KL_2}}= \mathbb{E}[D_{\mathrm{KL}}(p(z|c_k, bb_k)\|p(z|c_k))]$.
The objective of the layout generation module is:
\begin{equation}
L_{\mathrm{layout}}=\lambda_{\mathrm{recon}}L_{\mathrm{recon}}+\lambda_{\mathrm{KL_2}}L_{\mathrm{KL_2}}.
\end{equation}
The model is trained with teacher forcing where the ground truth bounding box at step $k$ will be used as the input for step $k+1$. 
At test time, the model will use the actual output boxes from previous steps. 
In addition,
the latent vector $z$ will be sampled from a conditional prior distribution $p(z|c_k)$, where $p$ is a prior encoder.

\vspace{\paramargin}
\paragraph{Bounding boxes with predefined shapes.}
In many design use cases, it is often required to constrain some design components to fixed size. 
For example, the logo size needs to be fixed in the ad design.
To achieve this goal, we augment the original layout generation module with an additional VAE encoder $\bar{h}_{bb}^{enc}$ to ensure the encoded latent vectors $z$ can be decoded to bounding boxes with desired widths and heights. 
Similar to \eqnref{stage2}, given a ground-truth bounding box $bb_k=(x_k,y_k,w_k,h_k)$, we obtain the reconstructed bounding box $\hat{bb}_k=(\hat{x}_k,\hat{y}_k,\hat{w}_k,\hat{h}_k)$ with $\bar{h}_{bb}^{enc}$ and $h_{bb}^{dec}$.
Then, instead of applying reconstruction loss on whole bounding boxes tuples, we only enforce the reconstruction of width and height with
\begin{equation}
    L^\mathrm{size}_\mathrm{recon} = \displaystyle\sum_{i=1}^{|O|}\lVert \hat{w}_i - w_i  \lVert_{1}+\lVert \hat{h}_i - h_i  \lVert_{1}.
\end{equation}
The objective of the augmented layout generation module is given by:
\begin{equation}
L'_{\mathrm{layout}}=\lambda^\mathrm{size}_{\mathrm{recon}}L^\mathrm{size}_{\mathrm{recon}}+L_{\mathrm{layout}}.
\end{equation}

\subsection{Layout Refinement}
\label{subsec:stage3}
\vspace{\subsecmargin}
We predict bounding boxes in an iterative manner that requires to fix the predicted bounding boxes from the previous iteration.
As a result, the overall bounding boxes might not be optimal, as shown in the layout generation module in \figref{framework}. 
To tackle this issue, we fine-tune the bounding boxes for better alignment and visual quality in the final layout refinement module.
Given a graph $G$ with ground-truth bounding boxes $\{bb_i\}$, we simulate the misalignment by randomly apply offsets $\delta\sim U(-0.05, 0.05)$ on $\{bb_i\}$, where $U$ is the uniform distribution.
We obtain misaligned bounding boxes $\{\bar{bb}_i\}$ = $\{bb_i+\delta_i\}$.
We apply a graph convolutional network $g_{\mathrm{ft}}$ for finetuning:
\begin{equation}
    \{\hat{bb}_i\} = g_{\mathrm{ft}}(G, \{\bar{bb}_i\}).
\end{equation}
The model is trained with reconstruction loss $L_{\mathrm{ft}}=\sum_i\lVert \{\hat{bb}_i\}-\{bb_i\}\lVert_{1}$.


\section{Experiments and Analysis}
\label{sec:experiment}
\vspace{\secmargin}

\paragraph{Datasets}
We perform the evaluation on three datasets: 
\begin{compactitem}
\item \tb{Magazine}~\cite{zheng2019contentaware}.
The dataset contains $4k$ images of magazine pages and $6$ categories (texts, images, headlines, over-image texts, over-image headlines, backgrounds).

\item \tb{RICO}~\cite{rico1,rico2}.
The original dataset contains $91k$ images of the Android apps interface and $27$ categories.
We choose $13$ most frequent categories (toolbars, images, texts, icons, buttons, inputs, list items, advertisements, pager indicators, web views, background images, drawers, modals) and filter the number of components within an image to be less than $10$, totaling $21k$ images.

\item \tb{Image banner ads}.
We collect $500$ image banner ads of the size $300\times250$ via image search using keywords such as ``car ads''.
We annotate bounding boxes of $6$ categories: images, regions of interest, logos, brand names, texts, and buttons. 
\end{compactitem}

\vspace{\paramargin}

\paragraph{Evaluated methods.}
We evaluate and compare the following algorithms:
\begin{compactitem}
\item \tb{sg2im}~\cite{johnson2018image}. 
 The model is proposed to generate a natural scene layout from a given scene graph.
The sg2im method takes as inputs graphs with complete relations in the setting where all constraints are provided.
When we compare with this method in the setting where no constraint is given, we simplify the input scene graph by removing all relations.
We refer the simplified model as \textbf{sg2im-none}.

\item \tb{LayoutVAE}~\cite{jyothi2019layoutvae}. 
This model takes a label set as input, and predicts the number of components for each label as well as the locations of each component.
We compare with the second stage of the LayoutVAE model (\ie the bounding box prediction stage) by giving the number of components for each label.
In addition, we refer to \textbf{LayoutVAE-loo} as the model that predicts the bounding box of a single component when all other components are provided and fixed (the leave-one-out setting). 

\item \tb{Neural Design Network}.  
We refer to \textbf{NDN-none} when the input contains no prior constraint,  \textbf{NDN-all} in the same setting as \textbf{sg2im} when all constraints are provided, and \textbf{NDN-loo} in the same setting as \textbf{LayoutVAE-loo}.
\end{compactitem}
We do not compare our method with LayoutGAN~\cite{li2019layoutgan} since LayoutGAN generates outputs in an unconditional manner (\ie generation from sampled noise vectors).
Even in the no-constraint setting, it is difficult to conduct fair comparisons as multiple times of resampling are required to generate the same combinations of components.

\subsection{Implementation Details}
In this work, $h_{bb}^{\mathrm{enc}}$, $h_{bb}^{\mathrm{dec}}$, and $h_{\mathrm{pred}}$ consists of $3$ fully-connected layers.
In addition, $g_c$, $g_p$, $g_\mathrm{enc}$, and $g_\mathrm{update}$ consist of $3$ graph convolution layers.
The dimension of latent vectors $z$ in the relation prediction and layout generation module is $32$.
The input features of nodes and edges are obtained from a dictionary mapping, which is trained along with the model.
For training, we use the Adam optimizer~\cite{adam} with batch size of $512$, learning rate of $0.0001$, and $(\beta_1, \beta_2) = (0.5, 0.999)$. 
In all experiments, we set the hyper-parameters as follows: $\lambda_{\mathrm{cls}}=1$,
$\lambda_{\mathrm{KL1}}=0.005$,  $\lambda_{\mathrm{recon}}=\lambda_{\mathrm{KL2}}=1$, and $\lambda_{\mathrm{recon}}=10$. 
We use a predefined order of component sets in all experiments. 

For the relation prediction module, the graphs with partial constraint are generated from the ground-truth graph with $0.2 \sim 0.9$ dropout rate. 
For the layout generation module, the input graphs with complete relations are constructed from the ground-truth layouts. 
The location and size relations are obtained by ground-truth bounding boxes. 
The corresponding outputs are the bounding boxes from the ground-truth layouts.

Since the location relations are discretized and mutually exclusive, there might be some ambiguity.
For example, a component is both ``above'' and ``right'' of another component when it is in the upper-right direction to the other.
To handle the ambiguity, we predefine the order when conflicts occur.
Specifically, ``above'' and ``below'' have higher priority than ``left of'' and ``right of''.

More implementation details can be found in the supplementary material.

\begin{table*}[t]
    \setlength{\tabcolsep}{2pt}
    \centering
    \scriptsize
    \caption{\textbf{Quantitative comparisons.} We compare the proposed method to other works on three datasets using three settings:
    no-constraint setting that no prior constraint is provided (first row), all-constraint setting that all relations are provided (second row), and leave-one-out setting that aims to predict the bounding box of a component with ground-truth bounding boxes of other components provided.
    The FID metric measures the realism and diversity, the alignment metric measures the alignment among components, and the prediction error metric measures the prediction accuracy in the leave-one-out setting. 
    }
    \begin{tabular}{@{}c cc cc cc} 
	    \toprule
		Datasets & \multicolumn{2}{c}{Ads}&\multicolumn{2}{c}{Magazine}&\multicolumn{2}{c}{RICO}
		\\  \cmidrule(lr){2-3} \cmidrule(lr){4-5} \cmidrule(lr){6-7}
		& FID $\downarrow$ & Align. $\downarrow$ & FID $\downarrow$ & Align. $\downarrow$ & FID $\downarrow$ & Align. $\downarrow$ \\
		sg2im-none & \textcolor{blue}{\textbf{116.63}} & \textcolor{blue}{\textbf{0.63}}& 95.81 & \textcolor{blue}{\textbf{0.97}} & 269.60 &\textcolor{blue}{\textbf{0.14}}\\
		LayoutVAE& 138.11$\pm38.91$& 1.21$\pm$0.08 & 81.56$\pm$36.78 &.314$\pm$0.11 &192.11$\pm$29.97 & 1.19$\pm$0.39\\
		NDN-none&  129.68$\pm$32.12 & 0.91$\pm$0.07 & \textcolor{blue}{\textbf{69.43$\pm$32.92}} & 2.51$\pm$0.09 & \textcolor{blue}{\textbf{143.51$\pm$22.36}}&0.91$\pm$0.03\\
		\midrule
		sg2im& 230.44 & 0.0069& 102.35 &0.0178 & 190.68 &0.007\\
		NDN-all& \textcolor{blue}{\textbf{168.44$\pm$21.83}} & \textcolor{blue}{\textbf{0.61$\pm$0.05}}&\textcolor{blue}{\textbf{82.77$\pm$16.24}} & \textcolor{blue}{\textbf{1.51$\pm$0.09}} &\textcolor{blue}{\textbf{64.78$\pm$11.60}} & \textcolor{blue}{\textbf{0.32$\pm$0.02}}\\
		\midrule\midrule
		 \cmidrule(lr){2-3} \cmidrule(lr){4-5} \cmidrule(lr){6-7}
		& Pred. error $\downarrow$ & Align. $\downarrow$ & Pred. error $\downarrow$ & Align. $\downarrow$ & Pred. error $\downarrow$ & Align. $\downarrow$ \\
		LayoutVAE-loo&0.071$\pm$0.002 &0.48$\pm$0.01 &0.059$\pm$0.002 & 1.41$\pm$0.02& 0.045$\pm$0.0021&0.39$\pm$0.02\\
		NDN-loo& \textcolor{blue}{\textbf{0.043$\pm$0.001}} &\textcolor{blue}{\textbf{0.36$\pm$0.01}} &\textcolor{blue}{\textbf{0.024$\pm$0.0002}} &\textcolor{blue}{\textbf{1.30$\pm$0.01}} & \textcolor{blue}{\textbf{0.018$\pm$0.002}}&\textcolor{blue}{\textbf{0.14$\pm$0.01}}\\
		\midrule
		real data& - & 0.0034 &- &0.0126 &- &0.0012\\
		\bottomrule
    \end{tabular}
    \label{tab:Quan}
    \vspace{\tabmargin}
\end{table*}

\vspace{\subsecmargin}
\subsection{Quantitative Evaluation}
\label{subsec:quan}
\noindent\textbf{Realism and accuracy.}
We evaluate the visual quality following Fr\'echet Inception Distance (FID)~\cite{fid} by measuring how close the distribution of generated layout is to the real ones.
We train a binary layout classifier to discriminate between good and bad layouts. The bad layouts are generated by randomly moving component locations of good layouts.
The classifier consists of four graph convolution layers and three fully connected layers.
The binary classifier achieves classification accuracy of $94\%$, $90$, and $95\%$ on the Ads, Magazine, and RICO datasets, respectively.
We extract the features of the second from the last fully connected layer to measure FID.

We measure FID in two settings.
First, a model predicts bounding boxes without any constraints.
That is, only the number and the category of components are provided.
We compare with LayoutVAE and sg2im-none in this setting.
Second, a model predicts bounding boxes with all constraints provided.
We compare with sg2im in this setting since LayoutVAE cannot take constraints as inputs.
The first two rows in \tabref{Quan} present the results of these two settings.
Since LayoutVAE and the proposed method are both stochastic models, we generate 100 samples for each testing design in each trial.
The results are averaged over $5$ trials.
In both no-constraint and all-constraint settings, the proposed method performs favorably against the other schemes. 

We also measure the prediction accuracy in the leave-one-out setting, \ie predicting the bounding box of a component when bounding boxes of other components are provided.
We measure the accuracy by the $L1$ error between the predicted and the ground-truth bounding boxes.
The third row of \tabref{Quan} shows the comparison to the LayoutVAE-loo method in this setting.
The proposed method gains better accuracy with statistical significance ($\ge$ 95\%), indicating that the graph-based framework encodes better relations among components.

\begin{table*}[t]
    \centering
    \caption{\textbf{Ablation on partial constraints and the refinement module.}
    We measure the FID and alignment of the proposed method taking different percentages of prior constraints as inputs using the RICO dataset. 
    We also show that the refinement module can further improve the visual quality as well as the alignment.}
        \begin{tabular}{@{}ccccc cc} 
	    \toprule
	    \begin{tabular}{@{}c@{}}Unary \\ size ($\%$)\end{tabular}  & 
	    \begin{tabular}{@{}c@{}}Binary \\ size ($\%$)\end{tabular} &
	    \begin{tabular}{@{}c@{}}Unary \\ location ($\%$)\end{tabular} &
	    \begin{tabular}{@{}c@{}}Binary \\ location ($\%$)\end{tabular}& Refinement & FID $\downarrow$ & Align. $\downarrow$\\
	    \midrule
	    0 & 0 & 0 & 0 & \cmark & 143.51$\pm$22.36&0.91$\pm$0.03 \\
	    \midrule
	    20 & 20 & 0 & 0 & \cmark &141.64$\pm$20.01&0.87$\pm$0.03\\
	    0 & 0 & 20 & 20 & \cmark &129.92$\pm$23.76 & 0.81$\pm$0.03\\
	    20 & 20 & 20 & 20 &  &126.18$\pm$23.11&0.74$\pm$0.02 \\
	    20 & 20 & 20 & 20 & \cmark & 125.41$\pm$21.68&0.70$\pm$0.02 \\
	    \midrule
	    100 & 100 & 100 & 100 &  & 70.55$\pm$12.68& 0.36$\pm$0.02\\
	    100 & 100 & 100 & 100 & \cmark & 64.78$\pm$11.60 & 0.32$\pm$0.02\\
	    \bottomrule
    \end{tabular}
    \label{tab:ablation}
    \vspace{\tabmargin}
\end{table*}

\begin{table}[t]
\begin{minipage}[t]{0.45\textwidth}

\caption{\textbf{Components Order.} 
We compare the performance of our model using different strategies of deciding orders of components.
We evaluate the FID score on the RICO dataset.}
\label{tab:order}
\centering
\begin{tabular}{c |ccc}
\toprule
Order & Size & Occurence & Random \\
\midrule
FID & 132.84 & 136.22 & 143.51 \\
\begin{tabular}{@{}c@{}}Pred. \\ error \end{tabular}& 1.08$\pm$0.04 & 1.02$\pm$0.04 & 0.91$\pm$0.03 \\
\bottomrule
\end{tabular}
\end{minipage}
\hfill
\begin{minipage}[t]{0.45\textwidth}

\caption{\textbf{Constraint consistency.}
We measure the consistency of the relations among generated components and the user-specified constraints.
}
\label{tab:consistency}

\centering
\begin{tabular}{c|ccc}
\toprule
Dataset & Ads & Magazine & RICO \\
\midrule
\begin{tabular}{@{}c@{}}Constraint \\ consistency (\%)\end{tabular}& 96.8 & 95.9 & 98.2 \\
\bottomrule
\end{tabular}
\end{minipage}
\end{table}


\noindent\textbf{Alignment.}
Alignment is an important principle in design creation. In most good designs, components need to be either in center alignment or edge alignment (e.g., left- or right-aligned). 
Therefore, in addition to realism, we explicitly measure the alignment among components using:
%
\begin{equation}
    \frac{1}{N_D}\displaystyle\sum_{d}\displaystyle\sum_{i}\mathop{\min}_{j, i\neq j} \{\min(l(c^d_i,c^d_j),m(c^d_i,c^d_j),r(c^d_i,c^d_j)\}),
\end{equation}
where $N_D$ is the number of generated layouts, $c^d_k$ is the $k_{th}$ component of the $d_{th}$ layout.
In addition, 
$l$, $m$, and $r$ are alignment functions where the distances between the left, center, and right of components are measured, respectively.

\tabref{Quan} presents the results in the no-constraint, all-constraint, and leave-one-out settings.
The results are also averaged over $5$ trials.
The proposed method performs favorably against other methods.
The sg2im-none method gets better alignment score since it tends to predict bounding boxes in several fixed locations when no prior constraint is provided, which leads to worse FID. 
For similar reasons, the sg2im method gains a slightly higher alignment score on RICO.

\noindent\textbf{Partial constraints.}
Previous experiments are conducted under the settings of either no constraints or all constraints provided.
Now, we demonstrate the efficacy of the proposed method on handling partial constraints.
%
%
\tabref{ablation} shows the results of layout prediction with different percentages of prior constraints provided.
We evaluate the partial constraints setting using the RICO dataset, which is the most difficult dataset in terms of diversity and complexity.
Ideally, the FID and alignment scores should be similar regardless of the percentage of constraints given.
However, in the challenging RICO dataset, the prior information of size and location still greatly improves the visual quality, as shown in \tabref{ablation},
The location constraints contribute to more improvement since they explicitly provide guidance from the ground-truth layouts.
As for the alignment score, layouts in all settings perform similarly.
Furthermore, the refinement module can slightly improve the alignment score as well as FID.

\vspace{\paramargin}
\noindent\textbf{User constraint consistency}
The major goal of the proposed model is to generate layouts according to user-specified constraints.
Therefore, we explicitly measure the consistency between the relations among generated components and the original user-specified constraints.
\tabref{consistency} shows that the generated layouts reasonably conform to the input constraints.

\begin{figure*}[t]
    \centering
    \includegraphics[width=\linewidth]{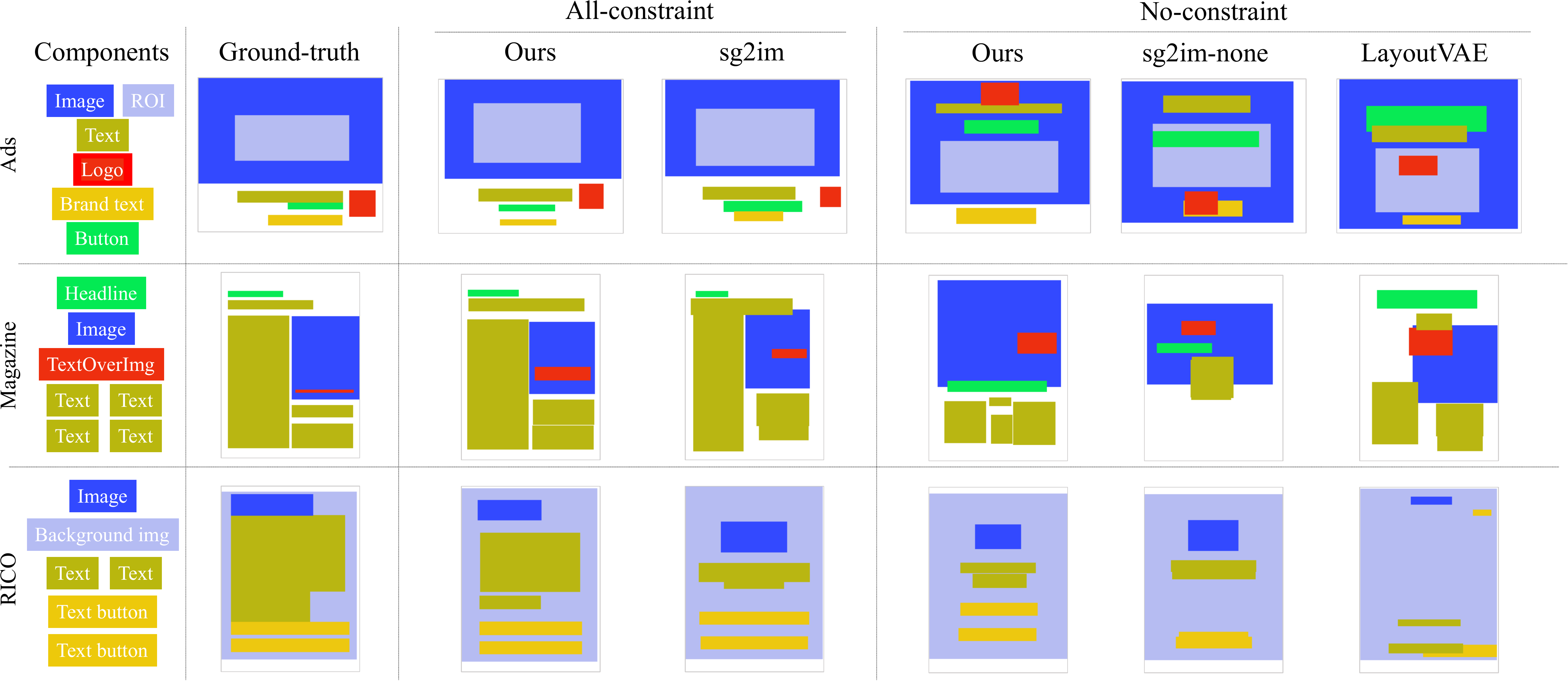}
    \caption{\textbf{Qualitative comparison.}
    We evaluate the proposed method with the LayoutVAE and Sg2im methods in both no-constraint and all-constraint setting.
    The proposed method can better model the relations among components and generate layouts of better visual quality.
    }
    \label{figure:qual}
    \vspace{\figmargin}
\end{figure*}

\vspace{\paramargin}
\noindent\textbf{Order of components.}
Since the proposed model predicts layouts in an iterative manner, the order of the components plays an important role.
We evaluate our method using three different strategies of defining orders: ordered by size, ordered by occurrences, and random order.
We show the comparisons in \tabref{order}.
We have a similar finding as in LayoutVAE that the order of components affects the generation results. 
However, we use the random order in all our experiments since our goal is not only to generate layouts, but also enable flexible user control.
In user cases such as leave-one-out prediction and layout recommendation, using random order can better align the training and testing scenarios.

\begin{figure*}[t]
    \centering
    \includegraphics[width=\linewidth]{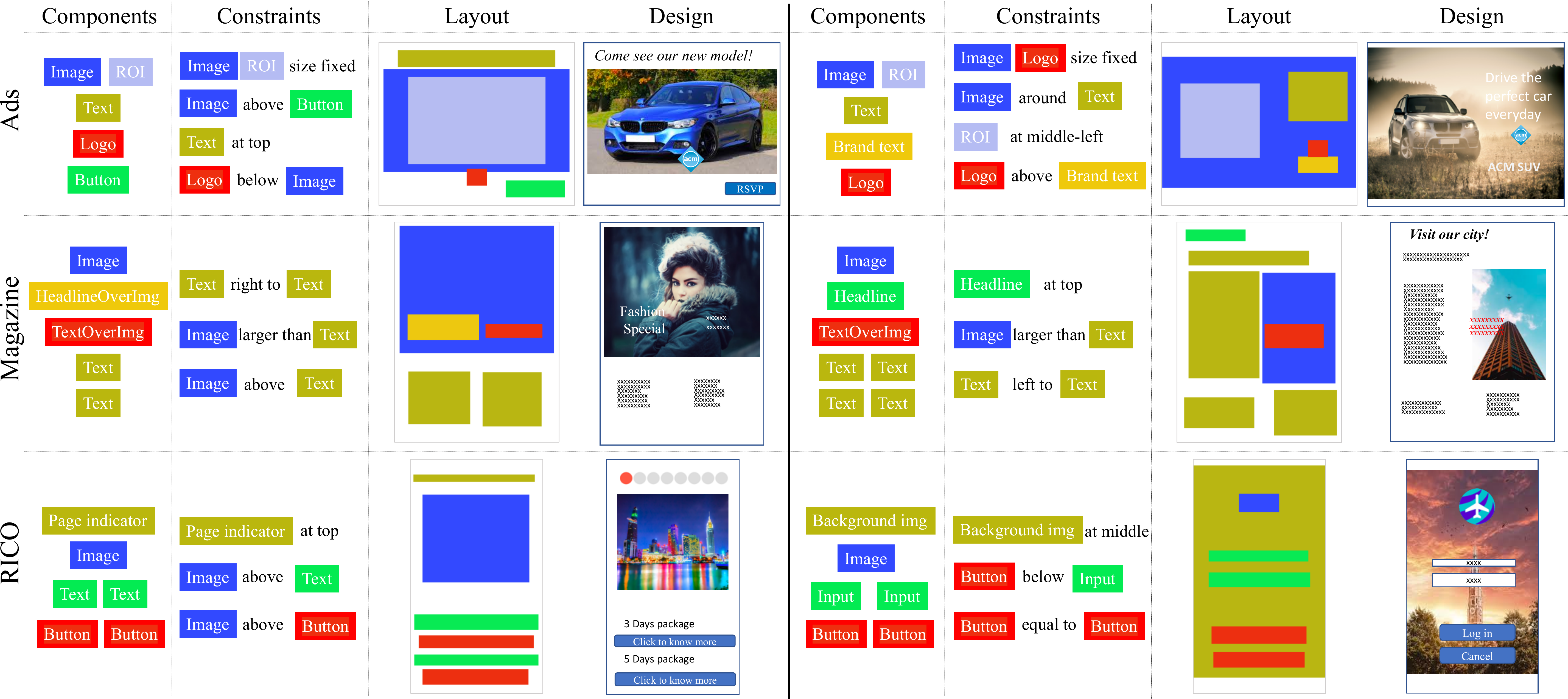}
    \caption{\textbf{Layout generation with partial user-specified constraints.}
    We generate layouts according to different user-specified constraints on location and size.
    Furthermore, we construct designs with real assets based on the generated layouts to better visualize the quality of our model.
    }
    \label{figure:examples}
    \vspace{\figmargin}
\end{figure*}
\vspace{\subsecmargin}
\subsection{Qualitative Evaluation}
\label{subsec:qual}
\vspace{\subsecmargin}
We compare the proposed method with related work in \figref{qual}.
In the all-constraint setting, both the sg2im method and the proposed model can generate reasonable layouts similar to the ground-truth layouts.
However, the proposed model can better tackle alignment and overlapping issues.
In the no-constraint setting, the sg2im-none method tends to place components of the same categories at the same location, like the ``text''s in the second row and the ``text''s and ``text button''s in the third row.
The LayoutVAE method, on the other hand, cannot handle relations among components well without using graphs. 
The proposed method can generate layouts with good visual quality, even with no constraint provided.

\vspace{\paramargin}
\noindent\textbf{Partial constraints.}
In \figref{examples}, we present the results of layout generation given several randomly selected constraints on size and location. 
Our model generates design layouts that are both realistic and follows user-specified constraints.
To better visualize the quality of the generated layouts, we present designs with real assets generated from the predicted layouts.
Furthermore, we can constrain the size of specific components to desired shapes (\eg we fix the \emph{image} and \emph{logo} size in the first row of \figref{examples}.) using the augmented layout generation module.

\begin{figure}[tb]
\begin{minipage}[b]{0.48\textwidth}
    \centering
    \includegraphics[width=\linewidth]{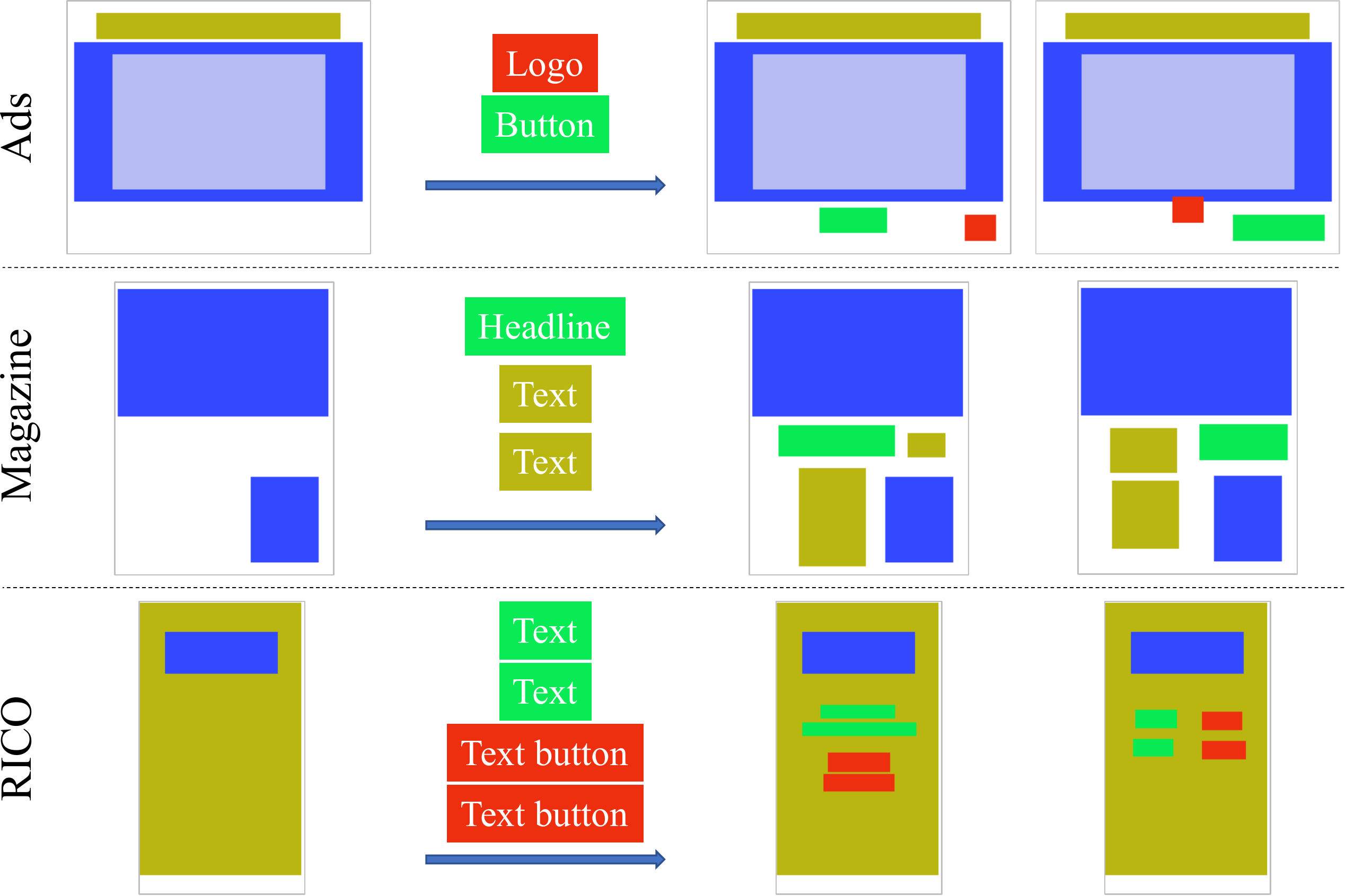}
    \captionof{figure}{\textbf{Layout Recommendation.} 
    We show examples of layout recommendations where locations of desired components are recommended given the current layouts.}
    
    \label{figure:app}
    \vspace{\figmargin}
\end{minipage}
\hfill
\begin{minipage}[b]{0.44\textwidth}
\centering
    \includegraphics[width=\linewidth]{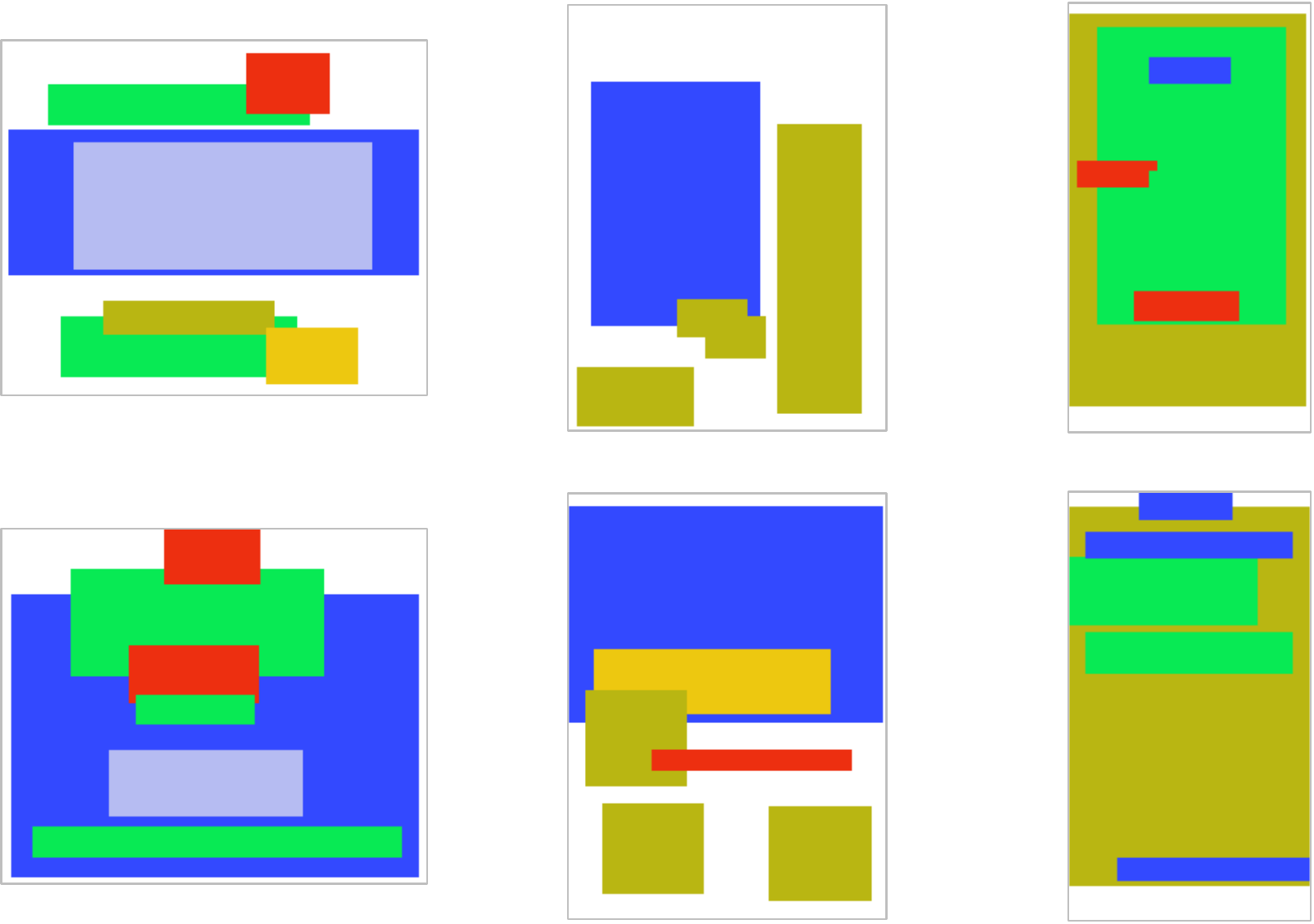}
    \captionof{figure}{\textbf{Failure cases.} 
    Generation may fail when the sampled latent vectors locate in under-sample spaces or the characteristics of inputs differ greatly from that in the training data.}
    \label{figure:failure}
    \vspace{\figmargin}
\end{minipage}

\end{figure}
\vspace{\paramargin}
\noindent\textbf{Layout recommendation.}
The proposed model can also help designers decide the best locations of a specific design component (e.g., \emph{logo}, \emph{button}, or \emph{headline}) when a partial design layout is provided. 
This can be done by building graphs with partial location and size relations based on the current canvas and set the relations to target components as \textit{unknown}.
We then complete this graph using the relation prediction module.
Finally, conditioned on the predicted graph as well as current canvas, we perform iterative bounding boxes prediction with the layout generation module.
\figref{app} shows examples of layout recommendations.

\vspace{\paramargin}
\noindent\textbf{Failure cases.}
Several reasons may lead to undesirable generation. 
First, due to the limited amount of training data, the sampled latent vectors used for generation might locate in undersampled spaces that are not fully exploited during training.
Second, the characteristic of the set of components is too different from the training data. 
For example, the lower-left image in \figref{failure} demonstrates a generation requiring three buttons and two logos, which are less likely to exist in real designs.
\section{Conclusion and Future Work}
\label{sec:conclusion}
In this work, we propose a neural design network to handle design layout generation given user-specified constraints.
The proposed method can generate layouts that are visually appealing and follow the constraints with a three-module framework, including a relation prediction module, a layout generation module, and a refinement module.
%
%
Extensive quantitative and qualitative experiments demonstrate the efficacy of the proposed model.
We also present examples of constructing real designs based on generated layouts, and an application of layout recommendation.

Visual design creation is an impactful but understudied topic in our community. It is extremely challenging. Our work is among one of the first works tackling graphic design in a well-defined setting that is reasonably close to the real use case. However, graphic design is a complicated process involving content attributes such as color, font, semantic labels, etc. Future directions may include content-aware graphic design or fine-grained layout generation beyond the bounding box.

\section{Acknowledgement}
This work is supported in part by the NSF CAREER Grant $\#1149783$.

\clearpage

\bibliographystyle{splncs04}
\bibliography{egbib}

\begin{thebibliography}{10}
\providecommand{\url}[1]{\texttt{#1}}
\providecommand{\urlprefix}{URL }
\providecommand{\doi}[1]{https://doi.org/#1}

\bibitem{bylinskii2017learning}
Bylinskii, Z., Kim, N.W., O'Donovan, P., Alsheikh, S., Madan, S., Pfister, H.,
  Durand, F., Russell, B., Hertzmann, A.: Learning visual importance for
  graphic designs and data visualizations. In: UIST (2017)

\bibitem{cheng2020rsegvae}
Cheng, Y.C., Lee, H.Y., Sun, M., Yang, M.H.: Controllable image synthesis via
  segvae. In: ECCV (2020)

\bibitem{damera2011probabilistic}
Damera-Venkata, N., Bento, J., O'Brien-Strain, E.: Probabilistic document model
  for automated document composition. In: DocEng (2011)

\bibitem{rico1}
Deka, B., Huang, Z., Franzen, C., Hibschman, J., Afergan, D., Li, Y., Nichols,
  J., Kumar, R.: Rico: A mobile app dataset for building data-driven design
  applications. In: UIST (2017)

\bibitem{duvenaud2015convolutional}
Duvenaud, D.K., Maclaurin, D., Iparraguirre, J., Bombarell, R., Hirzel, T.,
  Aspuru-Guzik, A., Adams, R.P.: Convolutional networks on graphs for learning
  molecular fingerprints. In: NeurIPS (2015)

\bibitem{goller1996learning}
Goller, C., Kuchler, A.: Learning task-dependent distributed representations by
  backpropagation through structure. In: ICNN (1996)

\bibitem{goodfellow2014generative}
Goodfellow, I., Pouget-Abadie, J., Mirza, M., Xu, B., Warde-Farley, D., Ozair,
  S., Courville, A., Bengio, Y.: Generative adversarial nets. In: NeurIPS
  (2014)

\bibitem{gori2005new}
Gori, M., Monfardini, G., Scarselli, F.: A new model for learning in graph
  domains. In: IJCNN (2005)

\bibitem{gupta2018imagine}
Gupta, T., Schwenk, D., Farhadi, A., Hoiem, D., Kembhavi, A.: Imagine this!
  scripts to compositions to videos. In: ECCV (2018)

\bibitem{fid}
Heusel, M., Ramsauer, H., Unterthiner, T., Nessler, B., Hochreiter, S.: {GANs}
  trained by a two time-scale update rule converge to a local nash equilibrium.
  In: NeurIPS (2017)

\bibitem{hong2018inferring}
Hong, S., Yang, D., Choi, J., Lee, H.: Inferring semantic layout for
  hierarchical text-to-image synthesis. In: CVPR (2018)

\bibitem{hurst2009review}
Hurst, N., Li, W., Marriott, K.: Review of automatic document formatting. In:
  DocEng (2009)

\bibitem{jain2016structural}
Jain, A., Zamir, A.R., Savarese, S., Saxena, A.: Structural-rnn: Deep learning
  on spatio-temporal graphs. In: CVPR (2016)

\bibitem{jin2018learning}
Jin, W., Yang, K., Barzilay, R., Jaakkola, T.: Learning multimodal
  graph-to-graph translation for molecular optimization. In: ICLR (2019)

\bibitem{johnson2018image}
Johnson, J., Gupta, A., Fei-Fei, L.: Image generation from scene graphs. In:
  CVPR (2018)

\bibitem{jyothi2019layoutvae}
Jyothi, A.A., Durand, T., He, J., Sigal, L., Mori, G.: Layoutvae: Stochastic
  scene layout generation from a label set. In: ICCV (2019)

\bibitem{karras2019style}
Karras, T., Laine, S., Aila, T.: A style-based generator architecture for
  generative adversarial networks. In: CVPR (2019)

\bibitem{adam}
Kingma, D., Ba, J.: Adam: A method for stochastic optimization. In: ICLR (2015)

\bibitem{kingma2013auto}
Kingma, D.P., Welling, M.: Auto-encoding variational bayes. In: ICLR (2014)

\bibitem{kipf2016semi}
Kipf, T.N., Welling, M.: Semi-supervised classification with graph
  convolutional networks. In: ICLR (2017)

\bibitem{kumar2011bricolage}
Kumar, R., Talton, J.O., Ahmad, S., Klemmer, S.R.: Bricolage: example-based
  retargeting for web design. In: SIGCHI (2011)

\bibitem{li2019layoutgan}
Li, J., Yang, J., Hertzmann, A., Zhang, J., Xu, T.: Layoutgan: Generating
  graphic layouts with wireframe discriminators. In: ICLR (2019)

\bibitem{li2019controllable}
Li, Y., Jiang, L., Yang, M.H.: Controllable and progressive image
  extrapolation. arXiv preprint arXiv:1912.11711  (2019)

\bibitem{rico2}
Liu, T.F., Craft, M., Situ, J., Yumer, E., Mech, R., Kumar, R.: Learning design
  semantics for mobile apps. In: UIST (2018)

\bibitem{o2014learning}
O’Donovan, P., Agarwala, A., Hertzmann, A.: Learning layouts for
  single-pagegraphic designs. TVCG  (2014)

\bibitem{pang2016directing}
Pang, X., Cao, Y., Lau, R.W., Chan, A.B.: Directing user attention via visual
  flow on web designs. ACM TOG (Proc. SIGGRAPH)  (2016)

\bibitem{razavi2019generating}
Razavi, A., Oord, A.v.d., Vinyals, O.: Generating diverse high-fidelity images
  with vq-vae-2. arXiv preprint arXiv:1906.00446  (2019)

\bibitem{rezende2014stochastic}
Rezende, D.J., Mohamed, S., Wierstra, D.: Stochastic backpropagation and
  approximate inference in deep generative models. In: ICML (2014)

\bibitem{scarselli2008graph}
Scarselli, F., Gori, M., Tsoi, A.C., Hagenbuchner, M., Monfardini, G.: The
  graph neural network model. TNN  (2008)

\bibitem{tabata2019automatic}
Tabata, S., Yoshihara, H., Maeda, H., Yokoyama, K.: Automatic layout generation
  for graphical design magazines. In: SIGGRAPH (2019)

\bibitem{tan2018text2scene}
Tan, F., Feng, S., Ordonez, V.: Text2scene: Generating abstract scenes from
  textual descriptions. In: CVPR (2019)

\bibitem{tseng2020retrievegan}
Tseng, H.Y., Lee, H.Y., Jiang, L., Yang, W., Yang, M.H.: Retrievegan: Image
  synthesis via differentiable patch retrieval. In: ECCV (2020)

\bibitem{zheng2019contentaware}
Xinru~Zheng, Xiaotian~Qiao, Y.C., Lau, R.W.: Content-aware generative modeling
  of graphic design layouts. SIGGRAPH  (2019)

\bibitem{yang2018graph}
Yang, J., Lu, J., Lee, S., Batra, D., Parikh, D.: Graph r-cnn for scene graph
  generation. In: ECCV (2018)

\bibitem{zhu2017toward}
Zhu, J.Y., Zhang, R., Pathak, D., Darrell, T., Efros, A.A., Wang, O.,
  Shechtman, E.: Toward multimodal image-to-image translation. In: NeurIPS
  (2017)

\end{thebibliography}
\end{document}